\documentclass[letterpaper, 10 pt, conference]{ieeeconf}  

\IEEEoverridecommandlockouts                              

\usepackage{graphicx}
\usepackage{epsfig} 
\usepackage{mathptmx} 
\usepackage{times} 
\usepackage{amsmath} 
\usepackage{amssymb}  
\usepackage{comment}
\usepackage[]{cleveref}   
\usepackage{booktabs}
\usepackage[caption=false,labelformat=empty]{subfig}
\usepackage{physics} 
\usepackage[symbol]{footmisc}

\title{\LARGE \bf
UncertaintyTrack: Exploiting Detection and Localization Uncertainty in Multi-Object Tracking
}

\author{Chang Won Lee and Steven L. Waslander
\thanks{Chang Won Lee and Steven L. Waslander  are with the Institute For Aerospace Studies (UTIAS), 
University of Toronto, Toronto, Canada. \newline
Emails: ({\tt\small john.lee@robotics.utias.utoronto.ca}; 
        {\tt\small steven.waslander@utoronto.ca})}
}

\newcommand{\baseline}{$^{*}$}

\begin{document}

\maketitle
\thispagestyle{empty}
\pagestyle{empty}


\begin{abstract}
Multi-object tracking (MOT) methods have seen a significant boost in performance recently, due to strong interest from the research community and steadily improving object detection methods. The majority of tracking methods, which follow the tracking-by-detection (TBD) paradigm, blindly trust the incoming detections with no sense of their associated localization uncertainty. This lack of uncertainty awareness poses a problem in safety-critical tasks such as autonomous driving where passengers could be put at risk due to erroneous detections that have propagated to downstream tasks, including MOT. While there are existing works in probabilistic object detection that predict the localization uncertainty around the boxes, no work in 2D MOT for autonomous driving has studied whether these estimates are meaningful enough to be leveraged effectively in object tracking. We introduce UncertaintyTrack, a collection of extensions that can be applied to multiple TBD trackers to account for localization uncertainty estimates from probabilistic object detectors. Experiments on the Berkeley Deep Drive MOT dataset show that the combination of our method and informative uncertainty estimates reduces the number of ID switches by around 19\% and improves mMOTA by 2-3\%. The source code is available at https://github.com/TRAILab/UncertaintyTrack
\end{abstract}

\section{INTRODUCTION}
Multi-object tracking (MOT) consists of tracking multiple dynamic objects by associating their detections across multiple frames. MOT for both image (2D) and lidar (3D) is a key component in robust autonomous driving systems as it provides object locations and scales across time for better scene understanding. MOT is also important for applications such as pedestrian counting and video analysis, which operate primarily in 2D. The object trajectories produced by MOT methods are used to predict future object motion, make safe decisions in motion planning for autonomous systems and identify human activities in surveillance settings. While MOT performance has improved significantly due to advances in object association, most state-of-the-art methods rely on accurate detections for strong performance.

The majority of MOT methods follow the tracking-by-detection (TBD) paradigm and separate detection and association into two sequential tasks, meaning the detections are provided as intermediate inputs to the association pipeline \cite{ramanan2003finding}. For this reason, TBD trackers are especially affected by detection performance. Unfortunately, despite their high accuracy on in-distribution data, deep object detectors are susceptible to errors with out-of-distribution (OOD) \cite{hendrycks2016OOD,lakshminarayanan2017proper_regression} examples often seen in deployment for real-world applications. For instance, the sources of error in autonomous driving could be due to complex, unforeseen environments such as night-time, snowy days, and object occlusions. These examples highlight the importance of not only developing robust and reliable autonomous systems to minimize the risks of failure, but also to provide outputs that express the uncertainty associated with object tracks so that OOD events can be approached with caution.

It is therefore critical to capture the uncertainty associated with MOT predictions to enable reliable integration of deep-learning-based systems into autonomy software stacks. By explicitly quantifying the underlying uncertainty of predictive outputs, developers can better understand where the network is uncertain, identify the sources of error, and use this information to make more informed decisions on the design for improvements. Additionally, an ideal robotic system should be able to flag predictions with high uncertainty and seek necessary human intervention autonomously \cite{harakeh2020bayesod}.

\begin{figure}[t]
\centering
\subfloat[]{%
    \centering
    \includegraphics[width=0.48\columnwidth]{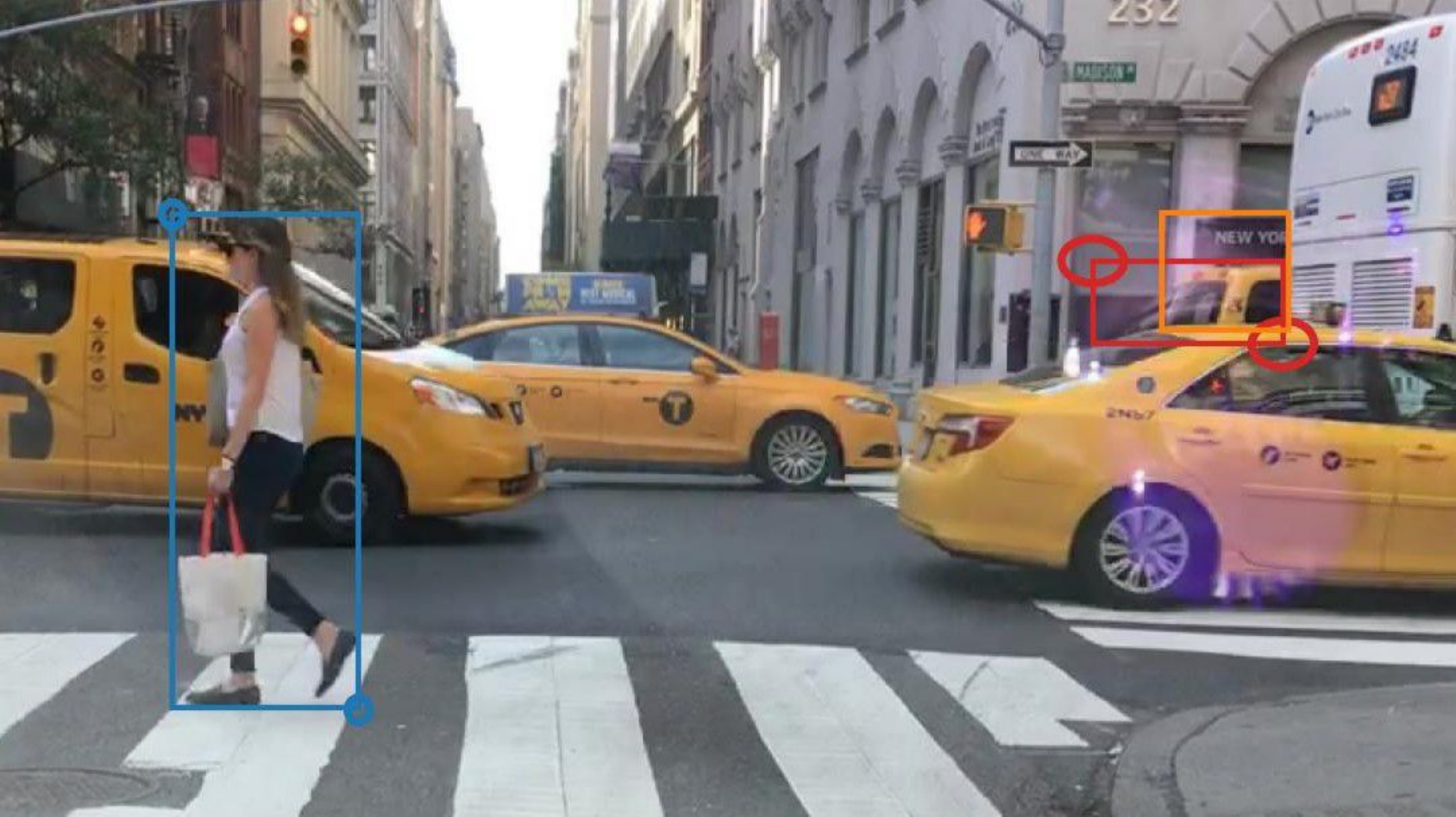}
    \hfill
    \includegraphics[width=0.48\columnwidth]{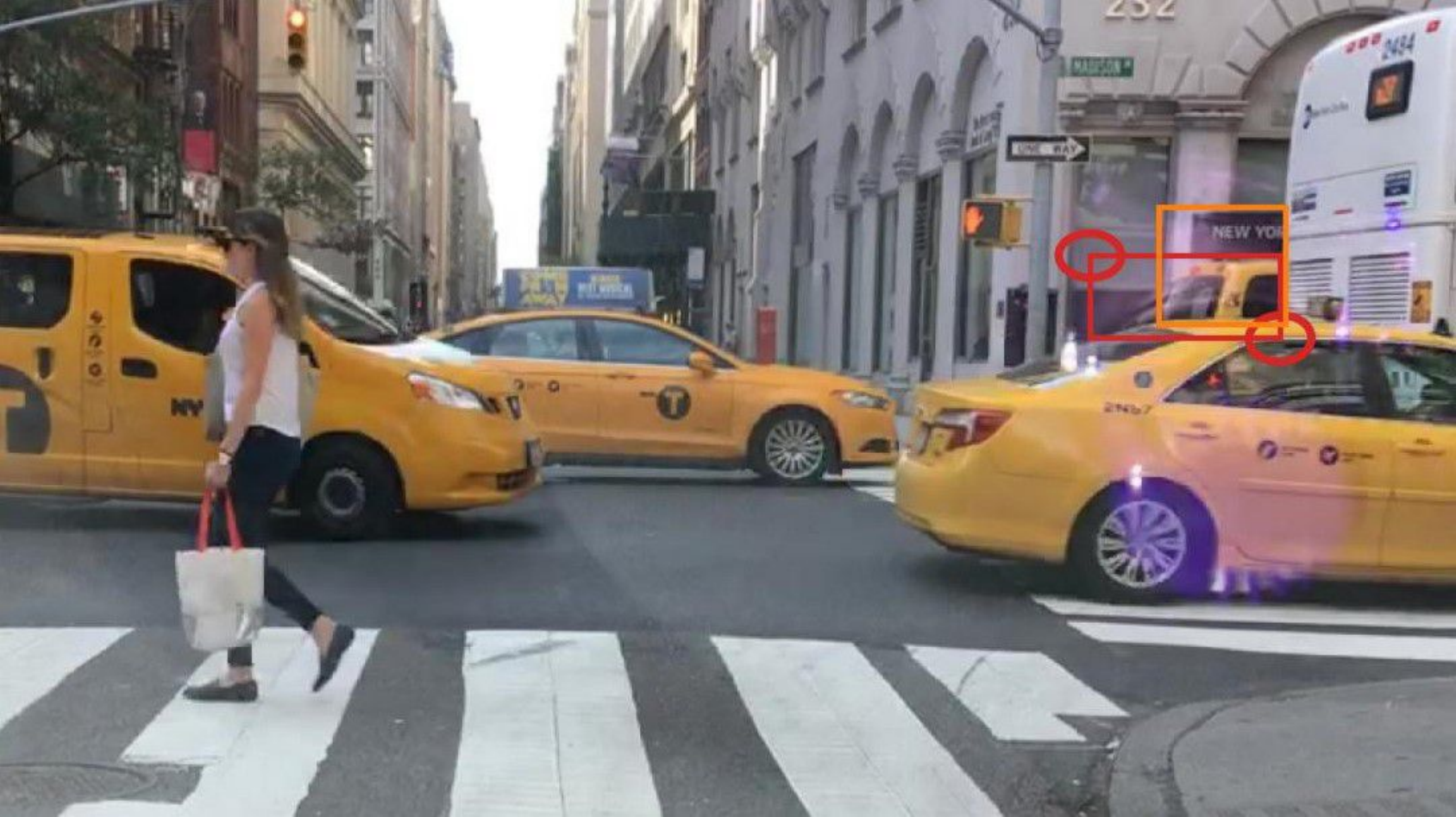}}
\vspace{-0.85\baselineskip}
\subfloat[]{%
    \centering
    \includegraphics[width=0.48\columnwidth]{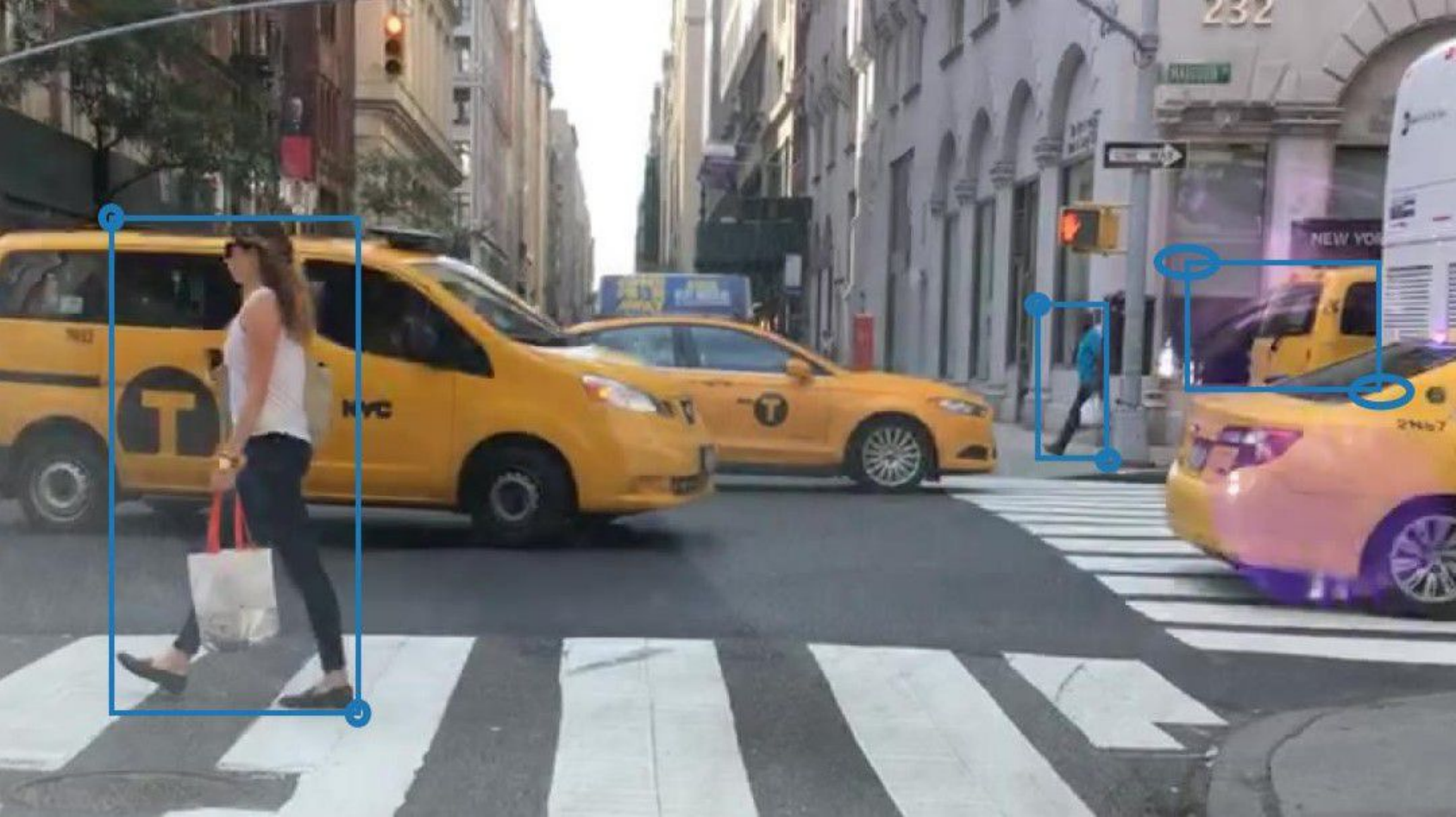}
    \hfill
    \includegraphics[width=0.48\columnwidth]{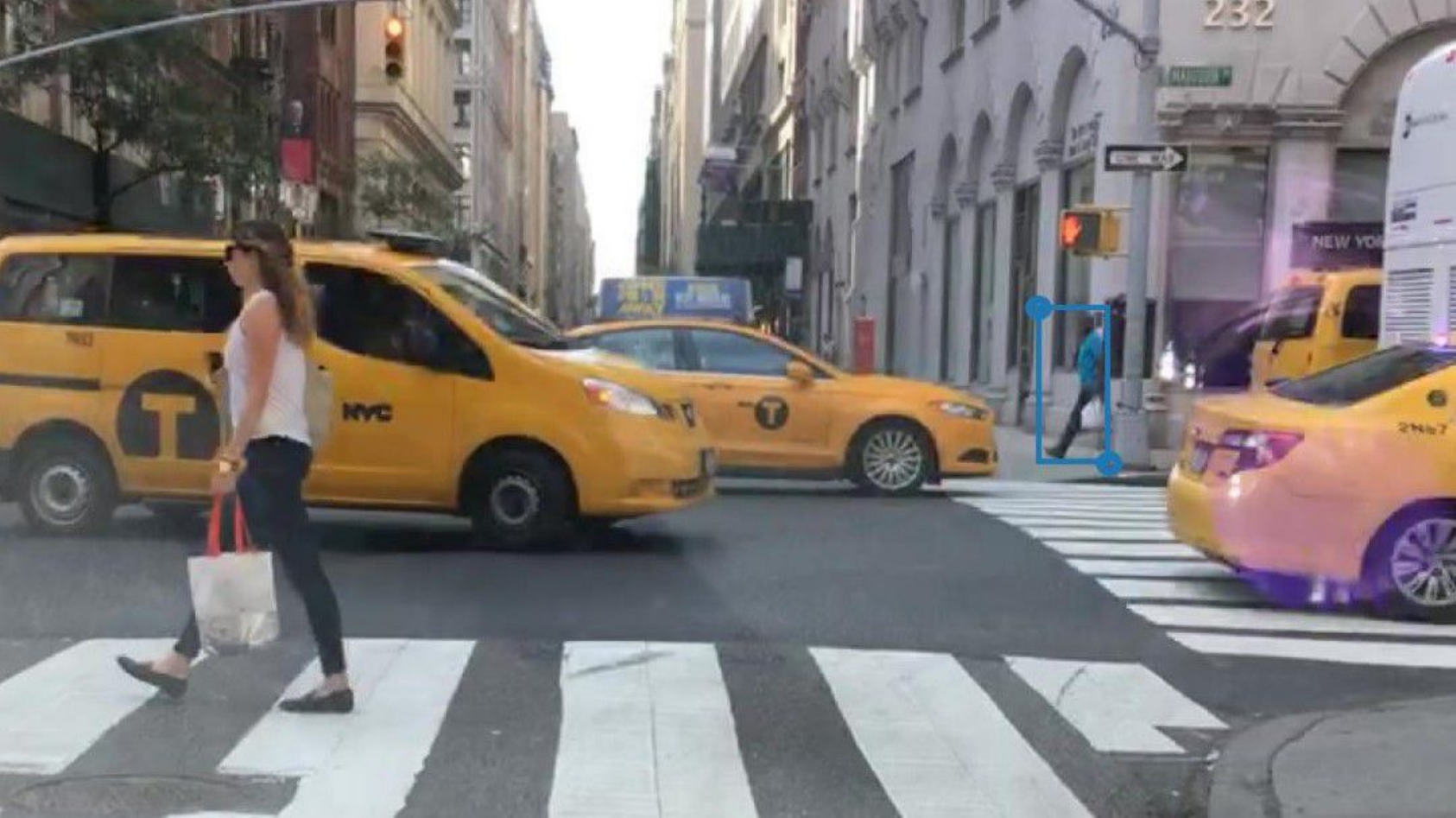}}
\vspace{-0.85\baselineskip}
\subfloat[]{%
    \centering
    \includegraphics[width=0.48\columnwidth]{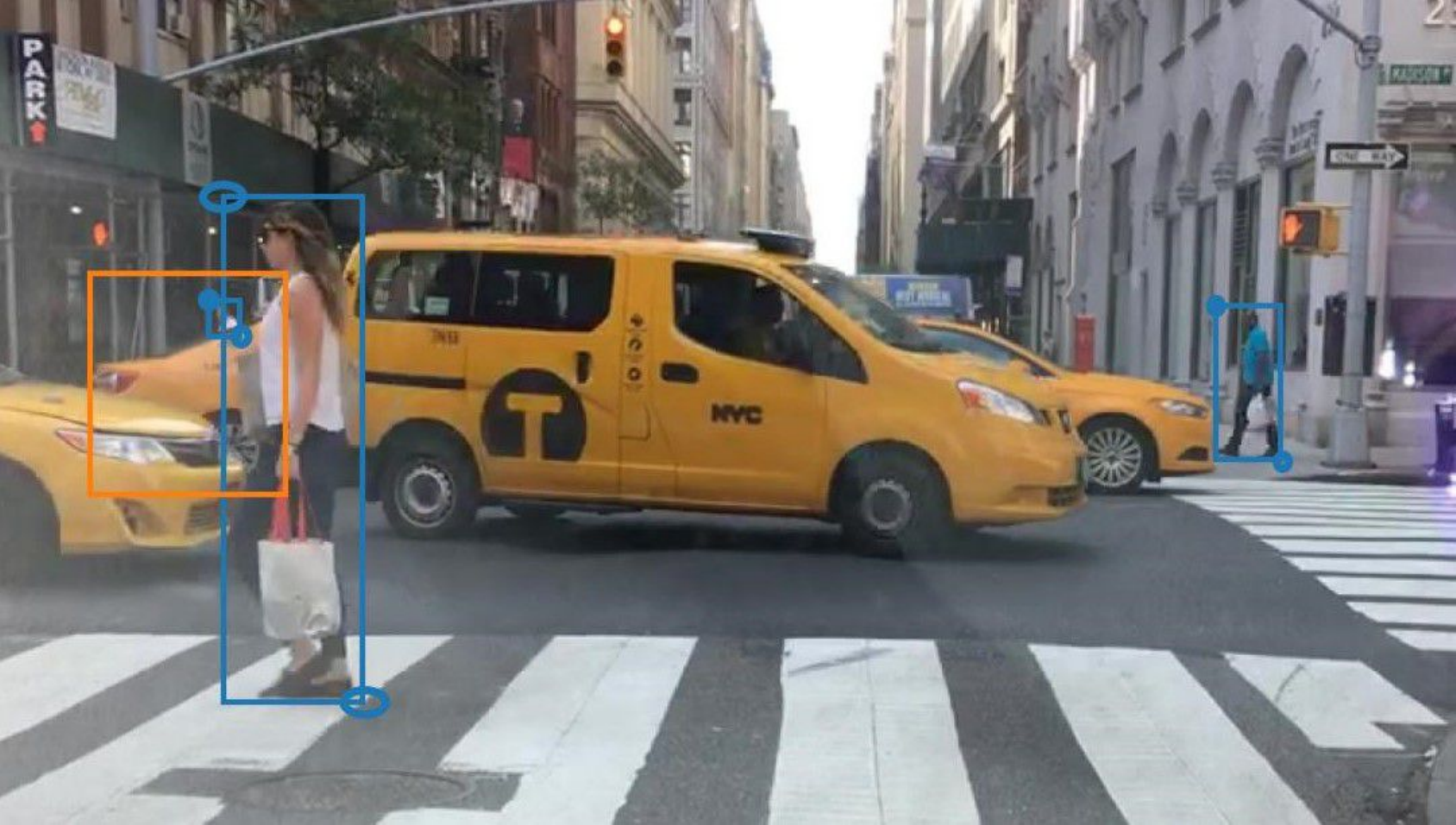}
    \hfill
    \includegraphics[width=0.48\columnwidth]{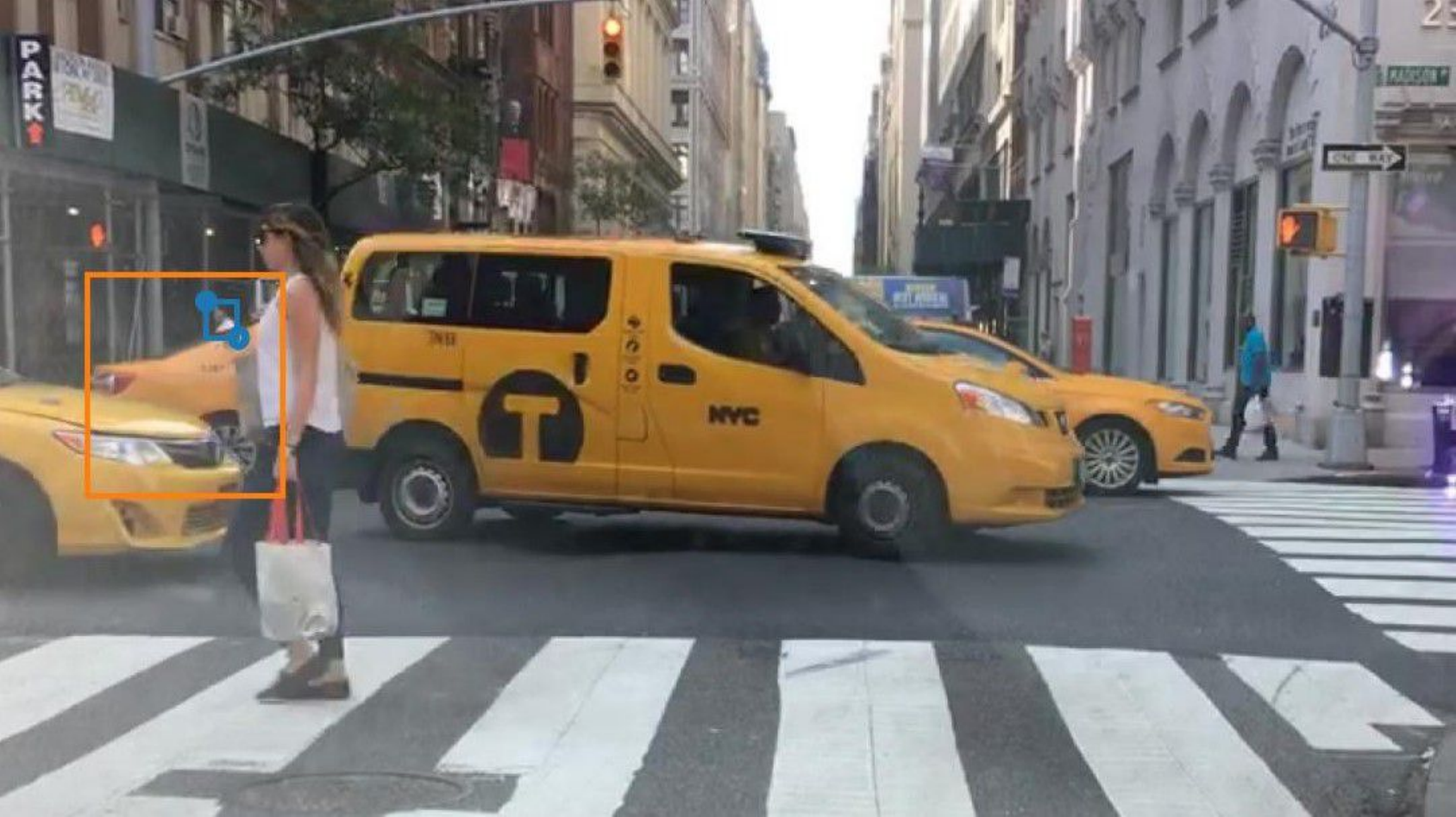}}
\vspace{-0.85\baselineskip}
\caption[]{
Examples of tracking results on the BDD100K dataset \cite{Yu_2020_BDD100}. In these images, orange boxes indicate false negatives, red boxes represent false positives, and blue boxes denote identity switches between tracked objects. The sequence of images is in chronological order, arranged from top to bottom.
    Compared to our baseline (left), UncertaintyTrack (right) effectively reduces the number of ID switches.}
\vspace{-1.0\baselineskip}
\label{fig:intro:uncertaintytrack_example}
\end{figure}

The introduction of Probabilistic Object Detection \cite{hall2020PDQ} has led to several probabilistic object detectors \cite{catak2021PURE, miller2019benchmarking, feng2019deep, le2018uncertainty, kraus2019onestageuncertainty} that quantify the semantic and spatial uncertainty of detections. However, there is no prior work in MOT that effectively exploits these uncertainty estimates to develop a more reliable tracking pipeline for autonomous driving. As current MOT methods are not designed to consider detection uncertainty from probabilistic object detectors when associating detections to tracks, their performance may suffer in the presence of high uncertainty. There are multiple ways trackers could benefit from such uncertainty estimates. For instance, they could filter out detections with high localization uncertainty or prioritize the association between confident predictions. To that end, we put forward the following contributions: 

\begin{itemize}
    \item We present Detection Uncertainty-Aware Kalman Filter, Confidence Ellipse Filtering, Bounding Box Relaxation and Entropy-Based Greedy Matching, which are simple and intuitive but effective extensions to existing 2D trackers to account for detection uncertainty. Our collection of add-ons that we denote UncertaintyTrack reduces the number of ID switches by around 19\% and increases mMOTA by 2-3\% on the Berkeley Deep Drive (BDD100K) dataset, when compared to existing baselines. 
    \item We study how the variation in uncertainty estimates due to the use of different datasets and detectors affects MOT performance.
    \item We show that the explicit parameterization of localization uncertainty distributions provides meaningful insight for human interpretability and identification of the sources of error in MOT.
\end{itemize}
The presented methods may also be extended to 3D methods with similar tracking pipelines, as there are no learned modules specific to 2D, such as appearance-based similarities.

\section{RELATED WORKS}
\subsection{Probabilistic Object Detection}

The majority of current state-of-the-art object detectors are composed of three primary building blocks: backbone, detection head, and the post-processing pipeline. The detection head maps high-level representations of the input data known as feature maps to object bounding boxes that express the location, orientation and extent of detected objects. Then, it is common to perform Non-Maximum Suppression (NMS) to filter out redundant boxes and return a single tight bounding box for each object. Detectors that follow this standard procedure \cite{ge2021yolox, ren2015faster, lin2017retina, duan2019centernet, redmon2018yolov3} are said to be deterministic because they output predicted bounding boxes with no estimates of the underlying uncertainty associated with the predictions.

In contrast, probabilistic object detectors are developed to produce reliable uncertainty estimations. Based on the Bayesian framework, predictive uncertainty can be divided into epistemic and aleatoric uncertainties. Epistemic uncertainties arise from model limitations that arise from training on a finite dataset, and aleatoric uncertainties are due to inherent noise in the data, such as far-away objects and occlusions \cite{kendall2017uncertainties}. Most probabilistic object detectors apply MC-Dropout \cite{gal2016dropout} or Deep Ensembles \cite{lakshminarayanan2017ensembles} at the detection head to model epistemic uncertainty and Direct Modeling to model aleatoric uncertainty \cite{feng2021review}. MC-Dropout involves estimating the uncertainty by computing sample statistics of detector outputs from multiple forward passes with dropout applied during inference, while ensembles use multiple copies of the network created with different training schemes to generate samples. Direct Modeling consists of using dedicated layers that learn to predict the parameters of output distributions. Direct Modeling relies on the assumption of a certain probability distribution over the predictions; categorical distributions are most common for classification tasks and the Gaussian is the most common for regression~\cite{feng2021review}.

To train the layers to learn bounding box distributions, it is common to use the negative log-likelihood (NLL) for Gaussian distributions \cite{kendall2017uncertainties}. 
However, it has been shown that minimizing the NLL to obtain the maximum likelihood estimates (MLE) of the network weights results in predicted distributions with high entropy (underconfident predictions). The energy score loss \cite{gneiting2008energy} for multivariate Gaussian was introduced as an alternative \cite{harakeh2021energy}:
\begin{equation} \label{related:eq:esloss}
    \begin{aligned}
    ES = \frac{1}{N}\sum_{n=1}^M\Bigl(&\frac1M\sum_{j=1}^{M}\|z_{n,j}-z_n\| \\
    &- \frac{1}{2(M-1)}\sum_{j=1}^{M-1}\|z_{n,j}-z_{n,j+1}\|\Bigr)
    \end{aligned}
\end{equation}
where there are $N$ ground truth bounding boxes $z_n$, $M$ i.i.d. samples drawn from $\mathcal{N}(\hat{\mu}(x_n,w),\Sigma(x_n,w))$ and $z_{n,j}$ is the $j$\textsuperscript{th} sample.

Finally, most probabilistic object detectors use NMS to follow their deterministic counterparts closely. Some others modify it to consider spatial uncertainty when removing redundancy \cite{feng2021review}. BayesOD, the basis for Prob-YOLOX in UncertaintyTrack, replaces it with Bayesian inference to consider all redundant outputs instead of discarding them \cite{harakeh2020bayesod}. Covariance Intersection \cite{smith2006cov_int} has also been proposed to fuse redundant detections with unknown correlation \cite{harakeh2021phd}.

\subsection{Multiple Object Tracking}

MOT methods that follow the tracking-by-detection (TBD) paradigm \cite{bewley2016SORT, cao2023ocsort, cetintas2023sushi, fischer2023qdtrack, zhang2022bytetrack, du2022strongsort} primarily focus on improving the re-identification of objects given their detections, as the object detection problem is already addressed by strong object detectors \cite{cai2018cascade, ge2021yolox, sun2021sparse}. For this reason, much of our focus will be on TBD methods to allow the use of bounding box distributions provided by probabilistic object detectors. 

The object association task is often set up as a bipartite graph matching problem where two sets of nodes in the graph represent the tracked objects and new observations, respectively. Existing works employ the Hungarian Algorithm \cite{kuhn1955hungarian} to find the optimal assignment between the two sets or use greedy matching \cite{zhou2020greedy}. SORT \cite{bewley2016SORT} uses a Kalman Filter that assumes constant velocity to predict the trajectories of tracked objects and measures the bounding box overlap with new detections which is used as similarity scores for the assignment problem. DeepSORT proposes to use the Mahalanobis distance between predicted Kalman object states and new detections after realizing that the IoU metric leads to a high number of ID switches when the motion model's state uncertainty is high \cite{wojke2017DeepSORT}. 

Furthermore, StrongSORT \cite{du2022strongsort} replaces the vanilla Kalman Filter with the NSA Kalman algorithm \cite{Du2021NSA} that adaptively calculates measurement noise covariance by taking into account the detection confidence score. The authors of OC-SORT \cite{cao2023ocsort} argue that the linear motion assumption leads to a large variance in estimated velocity by KF. They propose to decrease the error accumulation in KF by employing a smoothing strategy. Finally, ByteTrack \cite{zhang2022bytetrack} leverages confidence scores to create a secondary set of detections for more potential associations. However, these methods still use the IoU as their geometric distance metric.

In summary, most methods in 2D MOT for autonomous driving, if not all, have only used the state uncertainties from the Kalman Filter or class confidence scores as a measure of uncertainty in the tracking pipeline. Despite the importance of uncertainty, none of the previous works have taken into account localization uncertainty from probabilistic object detectors to show their effectiveness in MOT.

\section{LEVERAGING UNCERTAINTY IN MULTI-OBJECT TRACKING}
\subsection{Predicting Bounding Box Distributions}

\subsubsection{Prob-YOLOX Architecture}
We build upon YOLOX \cite{ge2021yolox} to develop Prob-YOLOX that estimates the localization uncertainty of the predicted bounding boxes. To explicitly parameterize the uncertainty distribution, we closely follow \cite{harakeh2020bayesod, harakeh2021phd} where the authors assume a multivariate Gaussian and predict the covariance matrix of the distribution. More specifically, we add a regression head that learns the log-scale diagonal covariance in parallel with the box regression head that predicts the mean. Then, the output of the two heads, the mean and covariance, parameterize a multivariate Gaussian that represents the bounding box distribution.

We emulate MC-Dropout sampling \cite{gal2016dropout} by computing the sample statistics of the bounding box delta samples drawn from the distribution as follows:
\begin{equation} \label{eq:sample_statistics}
\begin{aligned}
    \mathbf{\mu}_s &= \frac1N \sum_{i=1}^{N} \mathbf{x}_i \\
    \Sigma_s &= \frac{1}{N-1} \sum_{i=1}^{N} (\mathbf{x}_i - \mathbf{\mu}_s)(\mathbf{x}_i - \mathbf{\mu}_s)^\top 
\end{aligned}
\end{equation}
where $\mathbf{x}_i$ is the i-th bounding box sample. Then, to cluster bounding boxes together, we perform NMS to determine the cluster centers. However, instead of discarding the other cluster members, which are generally considered redundant detections, we employ the Improved Fast Covariance Intersection \cite{franken2005improved}, a tractable approximation of the Covariance Intersection algorithm \cite{smith2006cov_int} to fuse the correlated predictions, but whose correlation is unknown.

\subsubsection{Losses}
In YOLOX, the regression branch is trained using both the L1 Loss and the IoU Loss. To guide Prob-YOLOX to learn the parameters of the multivariate Gaussian distribution, we replace the L1 Loss with Energy Score Loss as proposed in \cite{harakeh2021energy}. Additionally, we introduce Sample-IoU Loss in place of the standard IoU Loss. This new loss function is designed to work in tandem with Energy Score Loss, helping to learn the distribution parameters while still optimizing the bounding box predictions to minimize the IoU, similar to how the IoU Loss supports the L1 Loss in YOLOX. The Sample-IoU is given by:
\vspace{-0.5\baselineskip}
\begin{equation} \label{method:eq:sampleiou}
    \begin{aligned}
    Sample-IoU = &\frac{1}{N}\sum_{n=1}^M\Bigl(\frac1M\sum_{j=1}^{M}IoU(z^{*}_{n,j}-z^{*}_n) \\
    &- \frac{1}{2(M-1)}\sum_{j=1}^{M-1}IoU(z^{*}_{n,j}-z^{*}_{n,j+1})\Bigr)
    \end{aligned}
\vspace{-0.5\baselineskip}
\end{equation}
where there are $N$ ground truth decoded bounding boxes \(z^{*}_n\), $M$ i.i.d. decoded samples drawn from \(\mathcal{N}(\hat{\mu}(x_n,w),\Sigma(x_n,w))\) and \(z^{*}_{n,j}\) is the $j$\textsuperscript{th} sample.

\subsection{Extending Existing Trackers To Leverage Uncertainty}
\begin{figure*}[t]
  \centering
  \includegraphics[width=0.8\textwidth]{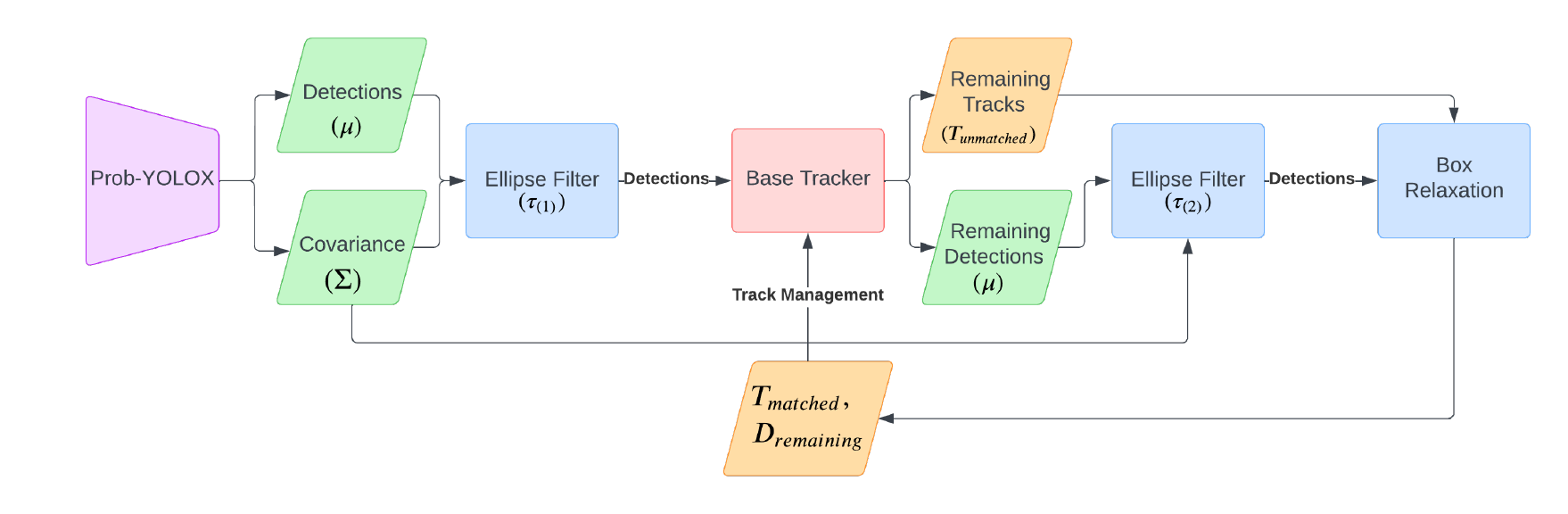}
  \caption{Overview of UncertaintyTrack. Prob-YOLOX (purple) outputs detections and their covariance matrices (green) which are initially filtered by Confidence Ellipse Filtering (blue). Following the base tracker, UncertaintyTrack applies the filter again before the Box Relaxation step. Matched tracks and remaining detections after this step are passed back to the tracker for track management.}
  \label{fig:method:UncertaintyTrack}
\vspace{-1.0\baselineskip}
\end{figure*}

We propose four possible ways to extend existing trackers to account for localization uncertainty in their association pipeline (see \cref{fig:method:UncertaintyTrack} for an overview of UncertaintyTrack). Our work builds upon tracking-by-detection algorithms that use the Kalman Filter as their motion model for two reasons: 1) the detector can be seamlessly replaced with a probabilistic object detector (e.g. Prob-YOLOX) that produces uncertainty estimates, and 2) the inclusion of an explicit motion model makes it more intuitive to incorporate detection uncertainty in the tracking task.

\subsubsection{Detection Uncertainty-Aware Kalman Filter}
The Kalman Filter (KF) \cite{kalman1960KF} acts as a motion predictor with a constant velocity assumption that models the object's location or state as:
\vspace{-0.5\baselineskip}
\begin{equation}
    \mathbf{x} = [x,y,a,h,\dot{x}, \dot{y}, \dot{a}, \dot{h}]^\top,
\label{eq:kf:state}
\vspace{-0.5\baselineskip}
\end{equation} 
representing the bounding box center position $(x,y)$, aspect ratio $a$, height $h$, and their respective velocities. To measure the geometric similarity between a tracked object and its potential observation, KF predicts the object's state at the next time step modelled as a multivariate Gaussian distribution (see \cite{barfoot2017state} for more details on KF). 

In current state-of-the-art MOT methods, the measurement uncertainty for KF is computed by scaling the aspect ratio with pre-determined coefficients that are inadaptive to changes in detection or track. In fact, this approach produces higher uncertainty for wider objects than taller ones and may cause issues in autonomous driving where both cars and pedestrians are present. Instead, we propose to replace this measurement uncertainty with the predicted covariance matrix from Prob-YOLOX.

For a single detection of an object, Prob-YOLOX outputs $\mathbf{\mu}_{box} = [x_1,y_1,x_2,y_2]^\top$, where $(x_1,y_1)$ and $(x_2,y_2)$ are the top-left and bottom-right corner pixel coordinates, and $\Sigma_{box} \in \mathbb{R}^{4\times4}$ that represents the bounding box covariance. Then, the predicted uncertainty can be incorporated into the KF update after it has been converted to the $[c_x, c_y, a, h]$ format.

\subsubsection{Confidence Ellipse Filtering}
The majority of tracking-by-detection algorithms filter incoming detections based on their class confidence score. By contrast, we suggest dropping detections with spatial uncertainty above a pre-determined threshold instead.  First, we define error ellipses for the top-left and bottom-right corners of each bounding box based on their 95\% confidence intervals. Then, we compute the ratio between their axis lengths and the box dimensions and compare it against a percentage threshold. Mathematically, if the box has width $w_{box}$ and height $h_{box}$ and the major and minor axis lengths of the error ellipse at the top-left and bottom-right corners are given by $a_{tl}$, $b_{tl}$, $a_{br}$ and $b_{br}$, respectively, only the detections with error ellipses that meet the following conditions are kept:
\vspace{-0.5\baselineskip}
\begin{equation} \label{eq:ellipse_filter}
\begin{aligned}
    \max(a_{tl}, a_{br}) &\leq \tau_{(i)} \times w_{box} \\
    &\& \\
    \max(b_{tl}, b_{br}) &\leq \tau_{(i)} \times h_{box}
\end{aligned}
\end{equation} 
where $\tau_{(i)}$ is a predefined percentage threshold.

\subsubsection{Bounding Box Relaxation}
After plotting the error ellipses along with the bounding boxes, we discovered that many ID switches are caused by an insufficient intersection between the detections and the tracks. Interestingly, for some detections that were incorrectly associated, their enlarged bounding boxes defined by their error ellipses had a significant overlap with the last observed detections of the tracks. Hence, we propose to add a final matching step with bounding boxes that have been enlarged according to the extremities of their error ellipses. 
More specifically, we compute $(x_{min}, y_{min})$ of the top-left error ellipse and the $(x_{max}, y_{max})$ of the bottom-right error ellipse. Then, the enlarged bounding box is defined by $[x_{min}, y_{min}, x_{max}, y_{max}]$. 

With this expansion, we forcefully increase the probability of the target being in the vicinity of the bounding box. We apply this uncertainty-based enlargement to both new detections and track detections at the previous step and match them using the GIoU \cite{rezatofighi2019giou} distance metric. In addition, to limit the amount of enlargement, we apply the ellipse-based filter a second time before this final matching step.

\subsubsection{Entropy-Based Greedy Matching}
Finally, at the final matching step with expanded boxes, we employ greedy matching based on the Gaussian entropy of the predicted box distributions to prioritize the association of detections with lower uncertainty (more confident predictions). 

\section{EXPERIMENTS AND RESULTS}
\subsection{Experimental Details}
\subsubsection{Datasets}

BDD100K \cite{Yu_2020_BDD100} is used as the primary benchmark for this study. The multi-class dataset contains scenes with diverse environments and varied weather conditions. We also evaluate UncertaintyTrack on the MOT17 pedestrian tracking dataset \cite{MOT17}. MOT17 does not have a separate validation set, so results are generated by taking the second half of the videos in the training set following \cite{zhou2020greedy}.

\subsubsection{Metrics}
\paragraph{Probabilistic Object Detection Metrics}

NLL and ES are used to evaluate the quality of predictive uncertainty. As proper scoring rules, their minimum score is obtained only when the predicted distribution matches the actual data generating distribution \cite{harakeh2021energy}. NLL is also a local scoring rule and penalizes low predictive densities at the exact values of the target. It also prefers predictive distributions that are less informative (high entropy). On the contrary, as a non-local measure, ES penalizes distributions with higher entropy and favors low entropy densities close to the target \cite{harakeh2021energy}, so the two metrics present complimentary assessments of the uncertainty prediction quality.

\paragraph{MOT Metrics}
We use the CLEAR MOT metrics \cite{bernardin2008clearmot} (e.g. MOTA, FP, FN, IDs), IDF1 \cite{ristani2016idf1} and HOTA \cite{luiten2021hota}. MOTA combines the number of false negatives (FN), false positives (FP), and ID switches (IDs) to measure the overall tracking performance, but it is biased toward detection performance. IDF1 computes the F1 score of the object trajectories but focuses on measuring association performance. HOTA seeks a balance between detection and association by taking the geometric mean of their accuracies. The `m' in front of the metrics denotes taking the average across all classes; for MOT17, it is equivalent to the original metrics.

\subsubsection{Implementation}

We leverage the open-source libraries MMDetection \cite{mmdetection} and MMTracking \cite{mmtrack2020} to develop Prob-YOLOX and UncertaintyTrack. We used the same setup as MMTracking's YOLOX setup for ByteTrack \cite{zhang2022bytetrack} to train Prob-YOLOX on MOT17 and BDD100K. 

We use ByteTrack and OC-SORT \cite{cao2023ocsort} as our baselines for UncertaintyTrack experiments as they are state-of-the-art methods on multiple MOT benchmarks and have been reproduced by MMTracking for smooth integration with custom object detectors. However, we remove linear interpolation from the post-processing pipeline to keep the methods online and relevant for robotics. We will use an asterisk $*$ to note that linear interpolation has been removed.

For inference, we use the same hyperparameters as the baselines for the base tracker. For $\tau_{(1)}$ and $\tau_{(2)}$, 0.65 and 0.3 were used for both ByteTrack\baseline and OC-SORT\baseline on BDD100K. For MOT17, (0.7, 0.3) and (0.6, 0.3) were used for ByteTrack\baseline and OC-SORT\baseline, respectively.
\subsection{Comparison With Existing Trackers} \label{sec:comparison_sota}

\begin{table}[t]
\begin{center}
\resizebox{\columnwidth}{!}{%
\begin{tabular}{@{}ccccccccc@{}}
\toprule
Method & Set & mMOTA$\uparrow$ & mIDF1$\uparrow$ & mHOTA$\uparrow$ & FP$\downarrow$ & FN$\downarrow$ & IDs$\downarrow$ & IDs (\%) \\ \midrule
ByteTrack\baseline  & BDD                & 32.5 & 42.1 & 38.4 & \textbf{21775} & 120104 & 46732 & 0.0\%   \\
\textbf{UncertaintyTrack (Ours)} & BDD & \textbf{35.1} & \textbf{45.5} & \textbf{40.1} & 27099 & \textbf{113488} & \textbf{37682} & \textbf{-19.4\%} \\ \midrule
ByteTrack\baseline     & MOT17 & 76.0 & 76.9 & 66.7 & 8114 & \textbf{29424} & 1231 & 0.0\%   \\
\textbf{UncertaintyTrack (Ours)} & MOT17 & \textbf{76.6} & \textbf{78.0} & \textbf{67.2} & \textbf{7218} & 29894 & \textbf{767} & \textbf{-37.7\%} \\ \midrule
OC-SORT\baseline         & BDD           & 27.0 & 39.4 & 36.8 & 40472 & \textbf{99363}  & 65814 & 0.0\%   \\ 
\textbf{UncertaintyTrack (Ours)} & BDD & \textbf{30.1} & \textbf{41.1} & \textbf{37.6} & \textbf{31898} & 108217 & \textbf{53217} & \textbf{-19.1\%} \\ \midrule
OC-SORT\baseline        &   MOT17          & 74.0 & 77.6 & 67.0 & 12099 & \textbf{27060}  & 2946 & 0.0\%   \\ 
\textbf{UncertaintyTrack (Ours)} & MOT17 & \textbf{75.8} & \textbf{78.9} & \textbf{67.5} & \textbf{7249} & 30511 & \textbf{1322} & \textbf{-55.1\%} \\ \bottomrule
\end{tabular}
}%
\end{center}
\caption{Comparison of ByteTrack\baseline and OC-SORT\baseline with UncertaintyTrack on BDD100K and MOT17 validation sets. The best results from each pair of comparisons are shown in bold. $*$ denotes the baseline without linear interpolation.}
\vspace{-1.0\baselineskip}
\label{results:table:mot:validation}
\end{table}

\begin{table}[t]
\begin{center}
\resizebox{\columnwidth}{!}{%
\begin{tabular}{@{}ccccccc@{}}
\toprule
Method & mMOTA$\uparrow$ & mIDF1$\uparrow$ & mHOTA$\uparrow$ & FP$\downarrow$ & FN$\downarrow$ & IDs$\downarrow$\\ \midrule
SUSHI \cite{cetintas2023sushi} & 40.3 & \textbf{60.0} & \textbf{48.4} & - & - & \textbf{13626} \\
QDTrack \cite{fischer2023qdtrack} & \textbf{42.4} & 55.6 & - & 89376 & \textbf{154797} & 14282 \\
ByteTrack \cite{zhang2022bytetrack} & 40.1 & 55.8 & - & 63869 & 169073 & 15466 \\
UncertaintyTrack$_{Byte}$ (Ours) & 33.5 & 47.6 & 40.9 & \textbf{52596} & 202856 & 73124 \\
UncertaintyTrack$_{OC}$ (Ours) & 29.0 & 42.4 & 38.1 & 61538 & 193916 & 102279 \\ \bottomrule
\end{tabular}
}%
\end{center}
\caption{Comparison between existing MOT methods and UncertaintyTrack on BDD100K test set. The best results are shown in bold.}
\vspace{-2.5\baselineskip}
\label{results:table:mot:bdd_test}
\end{table}

To address some of the shortcomings in MOT in the presence of detections with high uncertainty, we apply UncertaintyTrack to ByteTrack\baseline and OC-SORT\baseline and present the evaluation results on the BDD100K validation set and MOT17 half validation set in \cref{results:table:mot:validation}. Our proposed extensions for localization uncertainty improve the two trackers on mMOTA, mIDF1 and mHOTA and significantly reduce the number of ID switches on both datasets. While our primary objective is to show that detection uncertainty estimates can be helpful for understanding tracking failures and is not purely to develop a state-of-the-art tracker, we include our results on the BDD100K test set for completeness. It is important to note that we do not use learning-based affinity scores for association unlike existing methods, which can significantly boost performance on the BDD100K set.

\subsection{Relationship Between Detection Uncertainty \& Tracking}
\begin{figure}[t]
\centering
\subfloat[]{%
    \centering
    \includegraphics[width=0.48\columnwidth]{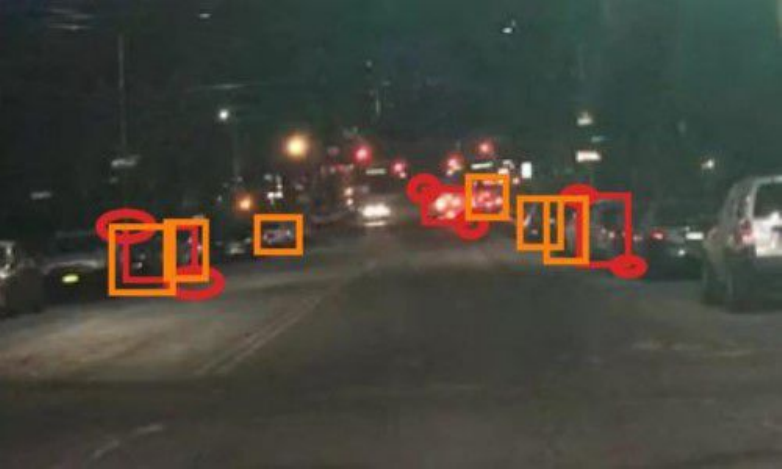}
    \hfill
    \includegraphics[width=0.48\columnwidth]{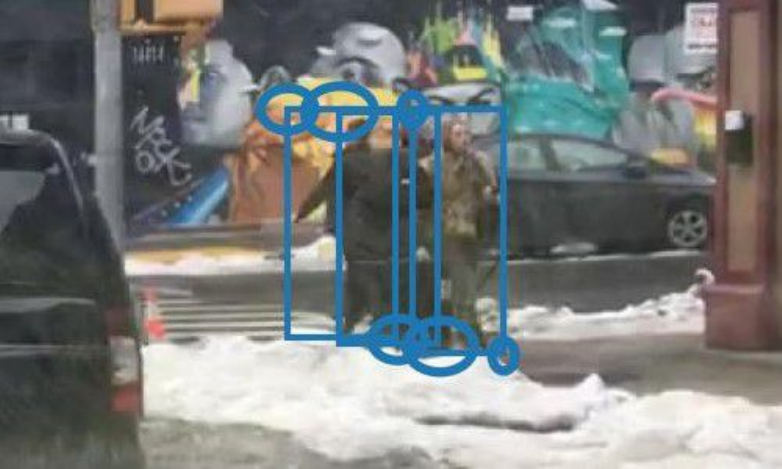}}
\vspace{-0.85\baselineskip}
\subfloat[]{%
    \centering
    \includegraphics[width=0.48\columnwidth]{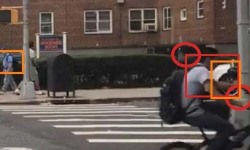}
    \includegraphics[width=0.48\columnwidth]{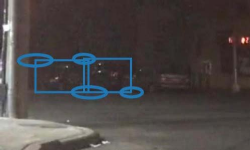}}
\vspace{-0.85\baselineskip}
\caption{
Examples of ByteTrack\baseline \cite{zhang2022bytetrack} failures on the BDD100K validation set.
Orange, red and blue represent false negatives, false positives and ID switches, respectively.}
\label{fig:results:error_visualization}
\end{figure}

\begin{table}[t]
\begin{center}
\resizebox{\columnwidth}{!}{%
\begin{tabular}{@{}ccccc@{}}
\toprule
Detector         & Set   & mAP (0.5:0.95)$\uparrow$ & NLL$\downarrow$ & ES$\downarrow$ \\ \midrule
YOLOX     & MOT17   &            60.5          &     -      & -    \\
YOLOX     & BDD     &            39.1          &     -    & -    \\ \midrule
Prob-YOLOX     & MOT17   &            60.0          &     12.6      & 9.81    \\
Prob-YOLOX     & BDD     &            39.0          &     639.76    & 9.61    \\
BayesOD & BDD     &          24.3            &     182.19    & 13.19 \\ \bottomrule
\end{tabular}
}%
\end{center}
\caption{Evaluation of detection and localization uncertainty quality from YOLOX \cite{ge2021yolox}, Prob-YOLOX and BayesOD \cite{harakeh2020bayesod}}
\vspace{-2.5\baselineskip}
\label{results:table:detection}
\end{table}

To understand why our proposed extensions were successful, it is necessary to analyze the causal relationship between inaccurate detections and tracking failures. We plot the 95\% confidence ellipses in each top-left and bottom-right corner of the detected bounding boxes that are false positives or lead to ID switches on the BDD100K validation set. No ellipses are drawn for false negative errors. In \cref{fig:results:error_visualization}, the two examples on the top show that some MOT errors are due to inaccurate detections of occluded/small objects, which aligns well with the expectations within the object detection community. Interestingly, they also manifest relatively large uncertainty ellipses, indicating the detector has learned to express the difficulty of the occluded/small object detection task well. In addition, other visible failures are caused by insufficient overlap between the detections and ground-truths (bottom-left) or a significant overlap between adjacent detections (bottom-right). Importantly, these errors also have high uncertainty associated with their boxes. 

With these few but informative illustrations, it is possible to infer that detections leading to MOT failures tend to have relatively high localization uncertainty compared to correctly associated detections. This highlights the importance of explicit parameterization of the uncertainty distributions as it allows for an in-depth analysis of where the ``black-box" deep object detectors are underperforming at inference time and a better understanding of how to adjust MOT association based on the distributions from probabilistic object detectors.                                                                                                                                            

Furthermore, to study why the improvements on BDD100K are more significant than what is observed on MOT17, we evaluate the quality of the predicted distributions of Prob-YOLOX on BDD100K and MOT17 using POD scoring rules and present our findings in \cref{results:table:detection}. Prob-YOLOX has a much higher NLL (local scoring rule) on BDD100K than MOT17, but lower ES (non-local scoring rule) and entropy scores, which signifies that the predicted distributions are informative and confident but centered away from the target values. In contrast, the distributions on MOT17 are closer to the target (lower NLL) while relatively having the same confidence level (similar ES). This observation aligns with the mAP values that show Prob-YOLOX has more accurate detections on MOT17 than on BDD100K. It may also explain UncertaintyTrack's behaviour on the two datasets. It is sensible that there is more room for improvement on the more complex BDD100K than MOT17 because the performance is saturating on the latter.

\subsection{Ablation Studies}
\subsubsection{BayesOD \& UncertaintyTrack}

\begin{table}[t]
\begin{center}
\resizebox{\columnwidth}{!}{%
\begin{tabular}{@{}ccccccccc@{}}
\toprule
Method       & Set      & mMOTA$\uparrow$ & mIDF1$\uparrow$ & FP$\downarrow$ & FN$\downarrow$ & IDs$\downarrow$ & IDs (\%) \\ \midrule
ByteTrack\baseline    & BDD          & \textbf{16.6} & \textbf{26.0} & \textbf{20638} & 236470 & 25336 & 0.0\% \\
\textbf{UncertaintyTrack (Ours)} & BDD & 12.1 & 25.7 & 66140 & \textbf{222100} & \textbf{23219} & \textbf{-8.4\%} \\ \midrule
ByteTrack\baseline  & MOT17   & \textbf{58.8} & \textbf{63.9} & \textbf{20470} & \textbf{43890} & 2205 & 0.0\%   \\
\textbf{UncertaintyTrack (Ours)} & MOT17 & 51.3 & 59.6 & 30350 & 46473 & \textbf{1847} & \textbf{-16.2\%} \\ \midrule
OC-SORT\baseline   & BDD             & -11.5 & 21.7 & 118483 & \textbf{144665}  & 89154 & 0.0\% \\ 
\textbf{UncertaintyTrack (Ours)} & BDD & \textbf{5.3} & \textbf{22.9} & \textbf{50277} & 200086 & \textbf{43217} & \textbf{-51.5\%} \\ \midrule
OC-SORT\baseline       & MOT17             & 40.3 & 54.6 & 53726 & \textbf{35398}  & 7439 & 0.0\%   \\ 
\textbf{UncertaintyTrack (Ours)} & MOT17 & \textbf{53.5} & \textbf{60.6} & \textbf{23812} & 30511 & \textbf{3298} & \textbf{-55.7\%} \\ \bottomrule
\end{tabular}
}%
\end{center}
\caption{Comparison of ByteTrack\baseline and OC-SORT\baseline with the extensions on BDD100K validation set and MOT17 validation set using BayesOD \cite{harakeh2020bayesod} as the detector. The best results from each pair of comparisons are shown in bold.}
\vspace{-2.5\baselineskip}
\label{results:table:mot:bayesod_validation}
\end{table}

To test if UncertaintyTrack could improve our baseline trackers with a different probabilistic detector, we repeated the same tracking experiments using BayesOD \cite{harakeh2020bayesod} and present the results in \cref{results:table:mot:bayesod_validation}. In contrast to earlier observations with Prob-YOLOX, the improvements are inconsistent among the trackers. While OC-SORT\baseline's performance generally increased on most metrics -- especially the number of ID switches, ByteTrack\baseline performed worse with UncertaintyTrack.

Similar to what was done for Prob-YOLOX, we evaluated BayesOD's uncertainty quality using the same POD metrics and included them in \cref{results:table:detection}. The results not only indicate BayesOD is a weaker detector with a lower mAP score, they show that the uncertainty estimates are less informative. BayesOD has a lower NLL but a higher ES score, meaning that the distributions are closer to the target but are not confident and more spread out. As UncertaintyTrack was designed to improve baseline trackers in the presence of informative and accurate predictive distributions, the limitations of the BayesOD lead to diminished performance in tracking. 

The observed discrepancy in UncertaintyTrack's results demonstrate that improvements in TBD trackers often depend on the detector choice and do not perform in the same way for all methods. It also further solidifies the importance of explicit parameterization of localization uncertainty for in-depth human analysis. Uncertainty estimates can be a complementary source of helpful information in designing a robust tracker as they allow for a better understanding of where detection errors cause MOT to fail.

\subsubsection{Effect of Each Component}
\begin{table}[t]
\begin{center}
\resizebox{\columnwidth}{!}{%
\begin{tabular}{@{}ccccccc@{}}
\toprule
KF Covariance & Ellipse & Relaxation & Greedy & mMOTA$\uparrow$ & mIDF1$\uparrow$ & IDs (\%)$\downarrow$ \\ \midrule
\checkmark & & & & 32.4 & 41.6 & -0.6\% \\
 & \checkmark & & & 32.6 & 42.1 & -4.1\% \\
  & \checkmark & \checkmark & & 34.8 & 45.1 & -14.3\% \\
  & \checkmark & \checkmark & \checkmark & 34.9 & 45.2 & -14.3\% \\
  \checkmark & \checkmark & \checkmark & \checkmark & \textbf{35.1} & \textbf{45.5} & \textbf{-19.4\%} \\ \bottomrule
\end{tabular}
}%
\end{center}
\caption{Analysis of the effect of UncertaintyTrack's components on ByteTrack\baseline. All versions are evaluated on the BDD100K validation set.}
\vspace{-2.5\baselineskip}
\label{results:table:ablation_studies}
\end{table}

We study the effect of each extension on ByteTrack\baseline and present the results on BDD100K validation set in \cref{results:table:ablation_studies}. The biggest improvement is observed with uncertainty-aware box relaxation, increasing mMOTA and mIDF1 by 2.2 and 3.0 points, respectively. Box relaxation in particular helps with missed associations between detections and tracks due to insufficient overlap. By increasing the size of the detection and tracklet boxes based on the corner uncertainty, their probability of intersection is essentially increased, leading to more matches.

There is a minimal increase in performance with ellipse-based filtering which hints that detections with high localization uncertainty were well-filtered by ByteTrack's class confidence-based filter and suggests an underlying correlation between the two. Also, the detections before the box relaxation step are generally far from each other for greedy matching to have sufficient impact.

Interestingly, the use of predicted box covariance as measurement uncertainty in the Kalman Filter complements the other components but decreases performance as a stand-alone extension. We hypothesize that this inconsistency in results may be due to the choice of distribution to model the uncertainty distribution. It is possible that the multivariate Gaussian distribution does not accurately represent the true underlying uncertainty distribution, suggesting that the uncertainty estimates might not be compatible with the Kalman Filter. This remains an area for future investigation.

\section{CONCLUSION}
This paper studies the quality of learned bounding box uncertainty from a probabilistic object detector and presents UncertaintyTrack, a collection of extensions applicable to several state-of-the-art tracking-by-detection MOT methods to leverage the predicted uncertainty. Experiments using UncertaintyTrack show that probabilistic object detectors produce meaningful localization uncertainty estimates which can be used in trackers to improve overall performance under complex scenarios. This work provided concrete examples that show the usefulness of probabilistic object detection for downstream tasks such as MOT and paves the path for more innovative probabilistic tracking techniques as well as more applications of probabilistic object detection. Potential future work includes extending UncertaintyTrack to 3D trackers and combining learned detection uncertainty with feature representation uncertainty in trackers that use learned similarities for association. 

\newpage
\bibliographystyle{IEEEtran}
\bibliography{main}

\end{document}